\documentclass[11pt,a4paper]{article}
\usepackage[hyperref]{emnlp2020}
\usepackage{times}
\usepackage{latexsym}

\usepackage{microtype}

\usepackage{subcaption}
\usepackage{graphicx}
\usepackage{mathtools}  
\usepackage{booktabs}
\usepackage{xspace}
\usepackage{xcolor}
\usepackage{comment}
\usepackage{soul}
\usepackage{url}

\aclfinalcopy 
\usepackage{enumitem}
\usepackage{arydshln}

\newcommand{\model}{Longformer\xspace}
\newcommand{\seqlen}{4,096\xspace}

\title{\model: The Long-Document Transformer}

\author{
Iz Beltagy\Thanks{\enspace Equal contribution.}
~~~~~~~~~Matthew E. Peters\textcolor{darkblue}{\footnotemark[1]}
~~~~~~~~~Arman Cohan\textcolor{darkblue}{\footnotemark[1]}\\
Allen Institute for Artificial Intelligence, Seattle, WA, USA\\
{\tt $\{$beltagy,matthewp,armanc$\}$@allenai.org}\\}

\date{}

\begin{document}
\maketitle
\begin{abstract}
Transformer-based models are unable to process long sequences due to their self-attention operation, which scales quadratically with the sequence length.
To address this limitation, we introduce the \model with an attention mechanism that scales linearly with sequence length, making it easy to process documents of thousands of tokens or longer.
\model's attention mechanism is a drop-in replacement for the standard self-attention and combines a local windowed attention with a task motivated global attention.
Following prior work on long-sequence transformers, 
we evaluate \model on character-level language modeling
and achieve state-of-the-art results on \texttt{text8} and \texttt{enwik8}.
In contrast to most prior work,
we also pretrain \model
and finetune it on a variety of downstream tasks.
Our pretrained \model consistently outperforms RoBERTa on long document tasks
and sets new state-of-the-art results on WikiHop and
TriviaQA. We finally introduce the Longformer-Encoder-Decoder (LED), a Longformer variant for supporting long document generative sequence-to-sequence tasks, and demonstrate its effectiveness on the arXiv summarization dataset.\footnote{\url{https://github.com/allenai/longformer}}

\end{abstract}

\section{Introduction}



Transformers~\cite{Vaswani2017AttentionIA} have achieved state-of-the-art results in a wide range of natural language tasks including generative language modeling~\cite{transformerxl,gpt2} and discriminative language understanding~\cite{bert}. 
This success is partly due to the self-attention component which enables the network to capture contextual information from the entire sequence. While powerful, the memory and computational requirements of self-attention grow quadratically with sequence length, making it infeasible (or very expensive) to process long sequences.

\begin{figure}[t!]
    \centering
    \includegraphics[width=1\linewidth]{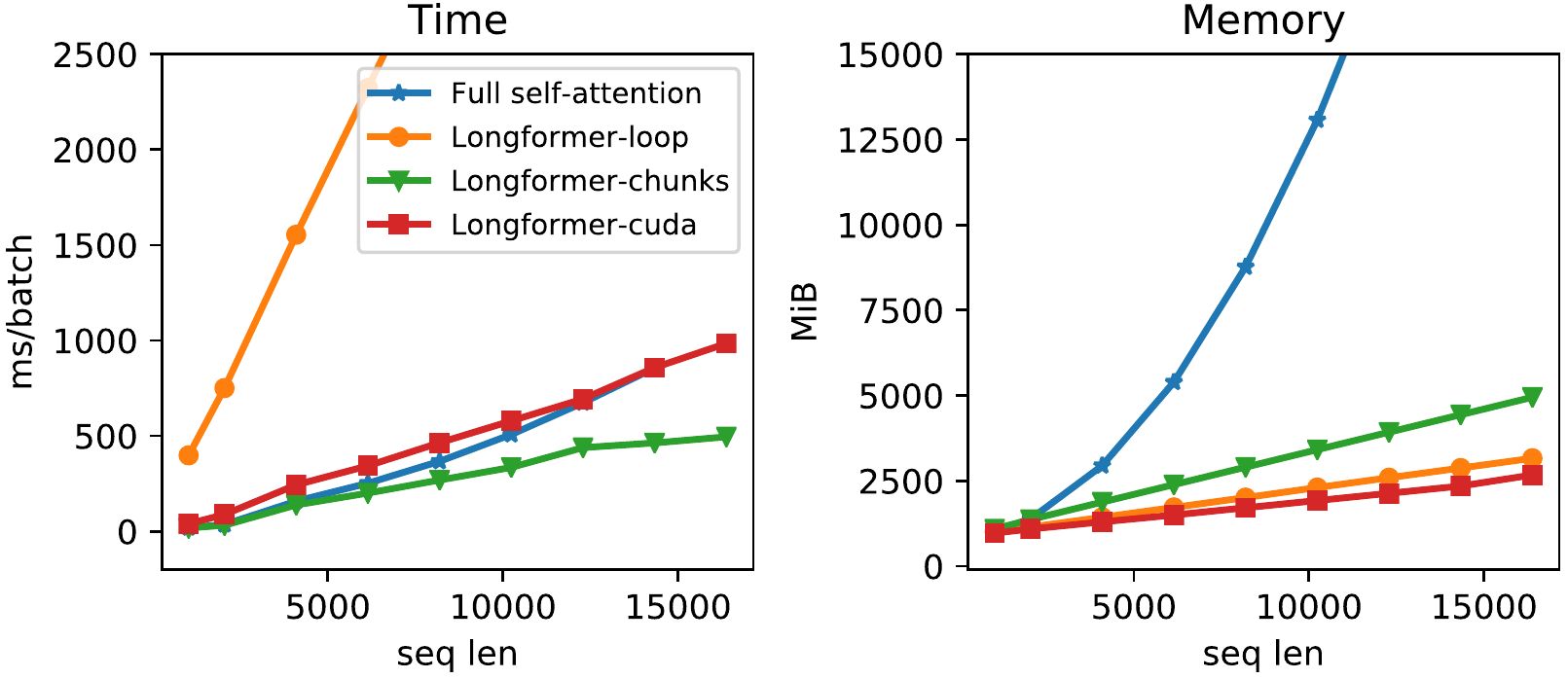}
    \caption{
Runtime and memory of full self-attention 
and different implementations of \model's self-attention; \texttt{\model-loop} is non-vectorized, \texttt{\model-chunk} is vectorized, and \texttt{\model-cuda} is a custom cuda kernel implementations.  
\model's memory usage scales linearly with the sequence length, unlike the full self-attention mechanism that runs out of memory for long sequences on current GPUs. Different implementations vary in speed, 
with the vectorized \texttt{\model-chunk} being the fastest. More details are in section~\ref{sec:tvm}.
    }
    \label{fig:tvm}
\end{figure}

To address this limitation, we present \model, a modified Transformer architecture
with a self-attention operation that scales linearly with the sequence length, making it versatile for processing long documents (Fig~\ref{fig:tvm}).
This is an advantage for natural language tasks such as long document classification, question answering (QA), and coreference resolution, where existing approaches partition or shorten the long context into smaller sequences that fall within the typical 512 token limit of BERT-style pretrained models. Such partitioning could potentially result in loss of important cross-partition information, and to mitigate this problem, existing methods often rely on complex architectures to address such interactions. On the other hand, our proposed \model is able to build contextual representations of the entire context using multiple layers of attention, reducing the need for task-specific architectures. 

Recent work has addressed the computational inefficiency of Transformers on long sequences (see Tab.~\ref{tab:related}).
However, they primarily focus on autoregressive language modeling (LM), 
while the application of long document transformers to document-level NLP tasks in the transfer learning setting \cite{NIPS2015_5949,Peters2018DeepCW,Howard2018UniversalLM,bert} has remained largely unexplored.  We address this gap and show that \model's attention mechanism can act as a drop-in replacement for the self-attention mechanism in pretrained Transformers, and leads to gains across a suite of document NLP tasks.

\model's attention mechanism is a combination of a windowed local-context self-attention and an end task motivated global attention that encodes inductive bias about the task.
Through ablations and controlled trials we show both attention types are essential -- the local attention is primarily used to build contextual representations, while the global attention allows \model to build full sequence representations for prediction.

We first evaluate \model on autoregressive character-level language modeling using a combination of windowed and a new dilated attention pattern, allowing the model to process sequences of up to 32K characters on modern GPUs. We achieve state-of-the-art results on \texttt{text8} and \texttt{enwik8} benchmark datasets, demonstrating the effectiveness of \model in long document modeling. 


Then, to evaluate \model's ability to replace the full self-attention operation of existing pretrained models, we pretrain
it with the masked language modeling (MLM) objective, continuing from the RoBERTa \cite{roberta} released checkpoint.
After pretraining, we apply it to downstream language tasks through finetuning and demonstrate that \model consistently outperforms RoBERTa on a wide range of document-level natural language tasks including text classification, QA, and coreference resolution, achieving state-of-the-art results on two of these datasets.

We finally introduce a variant of Longformer which instead of an encoder-only Transformer architecture, it follows an encoder-decoder architecture similar to the original Transformer model \cite{Vaswani2017AttentionIA}, and it is intended for sequence-to-sequence (seq2seq) learning \cite{Sutskever2014SequenceTS}. We call this model Longformer-Encoder-Decoder (LED) that uses Longformer's efficient attention pattern on the encoder network, allowing it to address long document seq2seq tasks such as summarization. We demonstrate the effectiveness of LED on the arXiv summarization dataset~\cite{arxiv2018}.

\begin{table}[t]
    \centering
    \small
    \setlength{\tabcolsep}{1pt}
    \begin{tabular}{@{}lcccr@{}}
    \toprule
    Model & attention & char-LM & other & pretrain  \\
          &    matrix       &         & tasks &   \\
    \midrule
    Transformer-XL~\shortcite{transformerxl} & ltr & yes & no & no \\
    Adaptive Span~\shortcite{adaptivespan} & ltr & yes & no & no \\
    Compressive~\shortcite{compressive} & ltr & yes & no & no \\ \hdashline[0.4pt/2pt]
    Reformer~\shortcite{reformer} & sparse & yes & no & no \\
    Sparse~\shortcite{sparseOpenai} & sparse & yes & no & no \\
    Routing~\shortcite{roy2020efficient} & sparse & yes & no & no \\
    \hdashline[0.4pt/2pt]
    BP-Transformer~\shortcite{BPTransformer} & sparse & yes & MT & no \\
    Blockwise~\shortcite{blockbert} & sparse & no & QA & yes \\
    \hdashline[0.4pt/2pt]
    Our \model & sparse & yes & multiple & yes \\
    \bottomrule
    \end{tabular}
    \caption{Summary of prior work on adapting Transformers for long documents. ltr: left-to-right.}
    \label{tab:related}
\end{table}

\begin{figure*}[t]
    \centering
    \begin{subfigure}[t]{0.22\textwidth}
        \centering
        \includegraphics[height=1.0in]{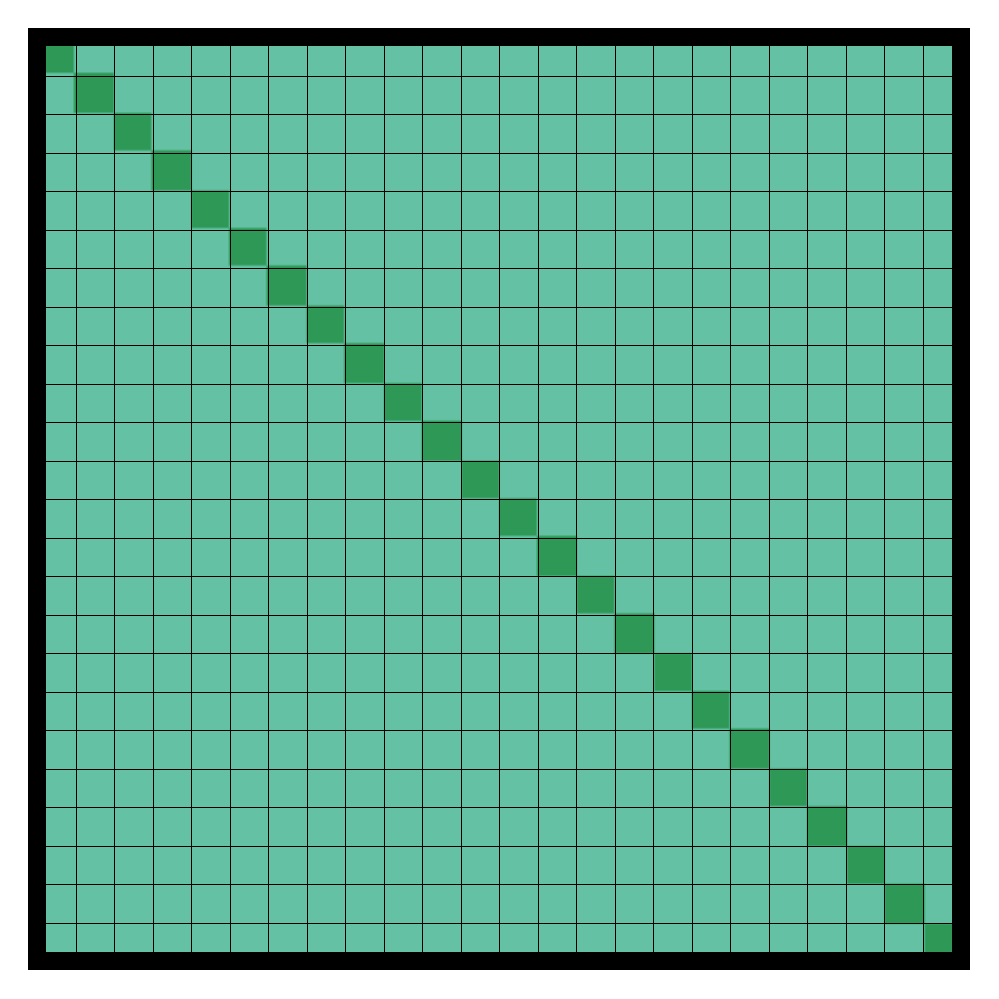}
        \caption{Full $n^2$ attention}
        \label{fig:attna}
    \end{subfigure}%
    ~~~~ 
    \begin{subfigure}[t]{0.22\textwidth}
        \centering
        \includegraphics[height=1.0in]{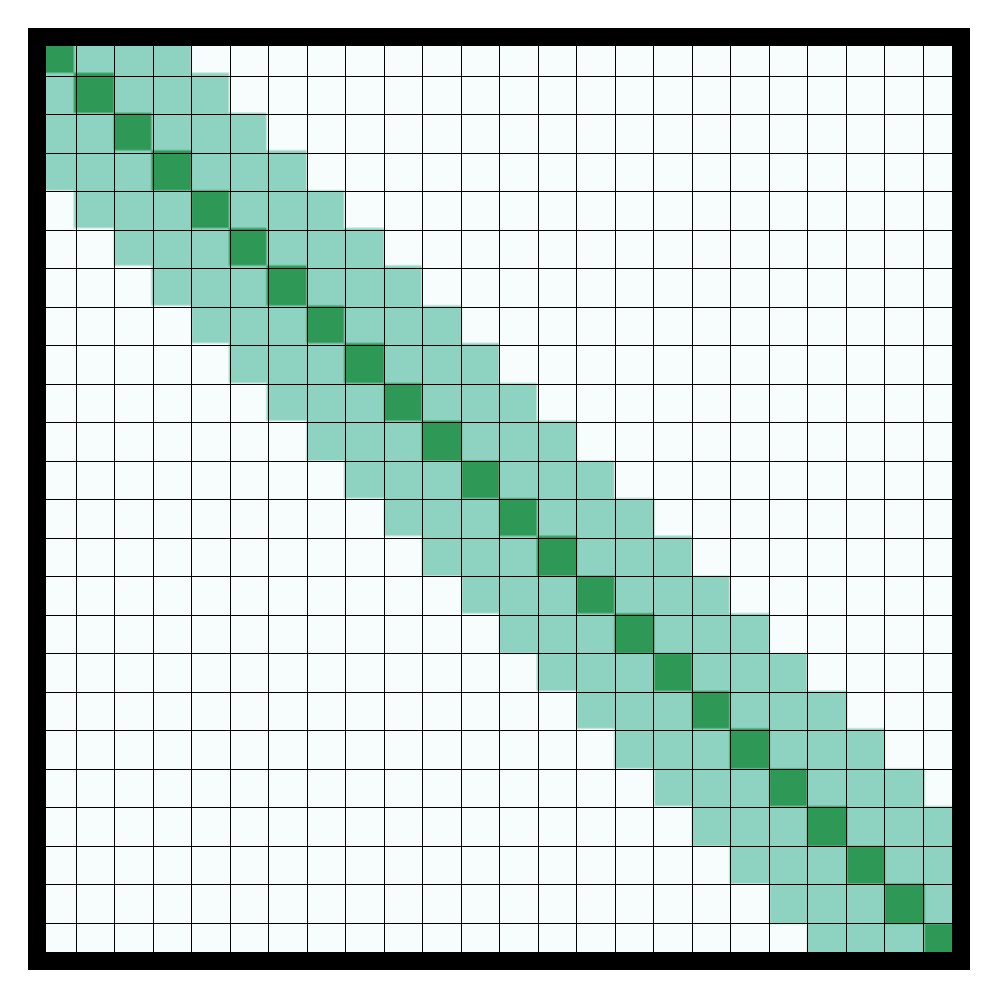}
        \caption{Sliding window attention}
        \label{fig:attnb}
    \end{subfigure}
    ~~~~
    \begin{subfigure}[t]{0.22\textwidth}
        \centering
        \includegraphics[height=1.0in]{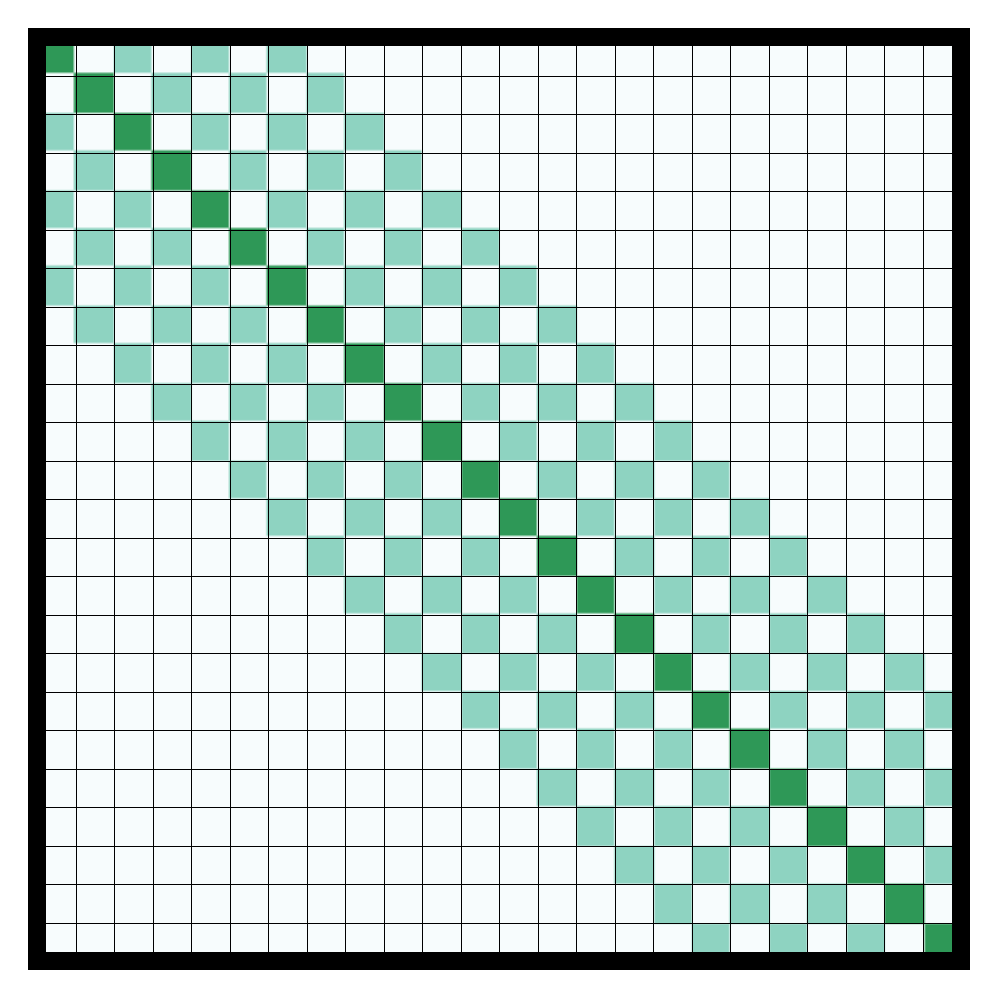}
        \caption{Dilated sliding window}
        \label{fig:attnc}
    \end{subfigure}
    ~~~~
    \begin{subfigure}[t]{0.22\textwidth}
        \centering
        \includegraphics[height=1.0in]{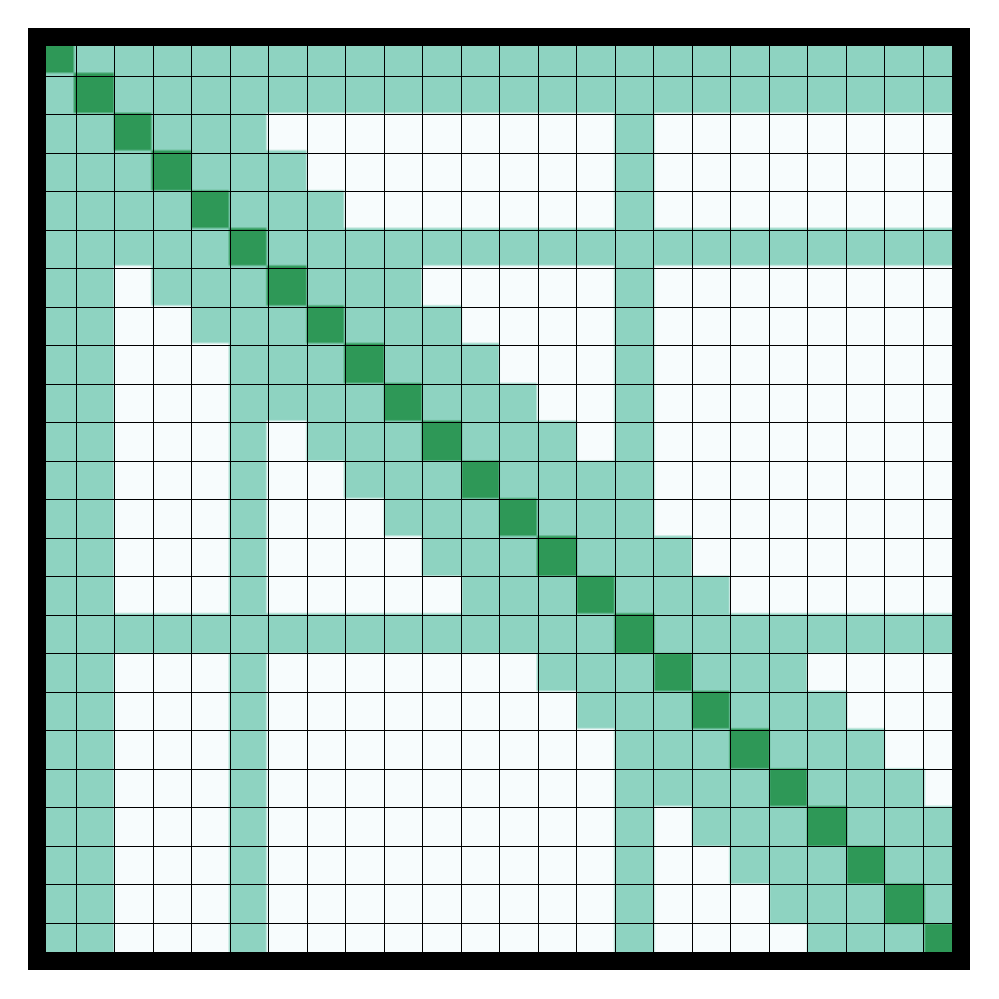}
        \caption{Global{+}sliding window}
        \label{fig:attnd}
    \end{subfigure}
    \caption{Comparing the full self-attention pattern and the configuration of attention patterns in our \model. 
    }
\label{fig:attn}
\end{figure*}

\section{Related Work}
\label{sec:related}

\paragraph{Long-Document Transformers}
Tab.~\ref{tab:related} summarizes recent prior work on long documents. Two types of self-attention approaches have been explored. 
The first is a left-to-right (ltr) approach that processes the document in chunks moving from left-to-right.
While such models have been successful in autoregressive language modeling, they are unsuitable for transfer learning approaches with tasks that benefit from bidirectional context.

Our work falls within the other general approach that defines some form of sparse attention pattern and avoids computing 
the full quadratic attention matrix multiplication.
The model with the most similar attention pattern to ours is Sparse Transformer~\cite{sparseOpenai}, which uses a form of dilated sliding window of blocks of size 8x8 provided by BlockSparse~\cite{blocksparse}. 
Our implementation (\S\ref{sec:model}) also includes a custom CUDA kernel, but it is more flexible and maintainable than BlockSparse which is implemented in C++, and designed for a specific version of TensorFlow.
We also introduce additional task motivated global attention patterns suitable for common NLP tasks
(\S\ref{sec:model}) and show they are essential for good performance in the transfer learning setting.


A few models tried tasks other than autoregressive language modeling,
which is 
a step forward because arguably focusing on 
language modeling as the primary evaluation
has led to the development of models with limited applicability.  BP-Transformer~\cite{BPTransformer} evaluated on 
machine translation (MT), but didn't explore the pretrain-finetune setting.
Blockwise attention~\cite{blockbert} pretrained their models and evaluated 
on question answering (QA). However, the evaluation is 
limited as it doesn't include language modeling, and the QA datasets
are of relatively short documents,\footnote{
SQuAD contexts typically fit within the 512 limit, 
and MRQA is constructed by dropping long-document examples. 
} therefore the effectiveness of this model on long document tasks remains unexplored.

\paragraph{Task-specific Models for Long Documents}
Many task-specific approaches have been developed to 
workaround the 512 limit of pretrained transformer models like BERT.
The simplest approach just truncates the document, commonly used for classification~\cite{truncateimdb}. 
Another approach chunks the document into chunks of 
length 512 (could be overlapping), processes each chunk separately, then combines the activations with a task specific model \cite{joshi-etal-2019-bert}.
A third approach popular for multihop and open domain QA tasks uses a two-stage model where the first stage retrieves relevant documents that are passed onto the second stage for answer extraction \cite{Clark2017SimpleAE,Chen2017ReadingWT}.
All of these approaches suffer from information loss due to truncation or cascading errors from the two stage approach.
In contrast, \model can process long sequences without truncating or chunking, allowing us to adopt a much simpler approach that concatenates the available context and processes it in a single pass. 

A few contemporaneous works\footnote{All were published on arXiv after Longformer.} have explored similar ideas to Longformer using local + global attention in Transformers, and pre-training it for long document natural language tasks. In particular, ETC \cite{ainslie-etal-2020-etc} uses a similar local + global attention instead of full self-attention to scale Transformers to long documents. Different from Longformer, ETC uses relative position embeddings (which we only used for the Autoregressive LM setting), introduces an additional training objective (CPC loss) for pre-training, and configures global attention in a slightly different way. It shows strong results on several tasks including reading comprehension and classification.
GMAT \cite{Gupta2020GMATGM} uses a similar idea of few global locations in the input serving as global memory. BigBird \cite{Zaheer2020BigBT} is an extension over ETC with evaluation on additional tasks, including summarization. Importantly, through theoretical analysis, BigBird shows that sparse Transformers are universal approximators of sequence functions and preserve these properties of the full self-attention.

\section{Longformer}
\label{sec:model}


The original Transformer model has a self-attention component with $O(n^2)$ time and memory complexity where $n$ is the input sequence length. 
To address this challenge, we sparsify the full self-attention matrix according to 
an ``attention pattern'' specifying pairs of input locations attending to one another.
Unlike the full self-attention, our proposed attention pattern scales linearly with the input sequence, making it efficient for longer sequences.
This section discusses the design and implementation of this attention pattern. 


\subsection{Attention Pattern}
\label{sec:attn}

\paragraph{Sliding Window}
Given the importance of local context \cite{Kovaleva2019RevealingTD}, our attention pattern employs a fixed-size window attention
surrounding each token.
Using multiple stacked layers of such windowed attention results in a large receptive field, where top layers have access to all input locations and have the capacity to build representations that incorporate information across the entire input, similar to CNNs \cite{Wu2019PayLA}.
Given a fixed window size $w$, each token attends to $\frac{1}{2}w$ tokens on each side (Fig.~\ref{fig:attnb}).
The computation complexity of this pattern is $O(n \times w)$, which scales linearly with input sequence length $n$. 
In a transformer with $\ell$ layers, the receptive field size at the top layer is $\ell \times w$ (assuming $w$ is fixed for all layers).
Depending on the application, it might be helpful to use different values of $w$ for each layer to balance between efficiency and model representation capacity (\S\ref{sec:charlm_attn}).

\paragraph{Dilated Sliding Window}
To further increase the receptive field without increasing computation, the sliding window can be ``dilated''. This is analogous to dilated CNNs~\cite{Oord2016WaveNetAG} where the window has gaps of size dilation $d$ (Fig.~\ref{fig:attnc}).
Assuming a fixed $d$ and $w$ for all layers, the receptive field 
is $\ell \times d \times w$, which can reach tens of thousands
of tokens even for small values of $d$. 

In multi-headed attention, each attention head computes a different attention score.  We found settings with different dilation configurations per head improves performance by allowing some heads without dilation to focus on local context, while others with dilation focus on longer context.


\paragraph{Global Attention} 

In state-of-the-art BERT-style models for natural language tasks, the optimal input representation differs from language modeling and varies by task.
For masked language modeling (MLM), the model uses local context to predict the masked word, while for classification, the model aggregates the representation of the whole sequence into a special token (\texttt{[CLS]} in case of BERT).
For QA, the question and document are concatenated, allowing the model to compare the question with the document through self-attention.

In our case, the windowed and dilated attention are not flexible enough to learn task-specific representations.
Accordingly, we add ``global attention'' on few pre-selected input locations.
Importantly, we make this attention operation symmetric: that is, a token
with a global attention attends to all tokens across the sequence, and all tokens in the sequence attend to it. 
Fig.~\ref{fig:attnd} shows an example of a sliding window attention
with global attention at a few tokens at custom locations. For example for classification, global attention is used for the \texttt{[CLS]} token while in QA global attention is provided on all question tokens.
Since the number of such tokens is small relative to and independent of $n$
the complexity of the combined local and global attention is still $O(n)$.
While specifying global attention is task specific, it is a easy way to add inductive bias to the model's attention, and it is much simpler than existing task specific approaches that use complex architecture to combine information across smaller input chunks.



\paragraph{Linear Projections for Global Attention}
Recall that given the linear projections $Q$, $K$, $V$, the Transformer
model \cite{Vaswani2017AttentionIA} computes attention scores as follows:
\begin{align}
\label{eq:qkv}
\text {Attention}(Q, K, V)=\operatorname{softmax}\left(\frac{Q K^{T}}{\sqrt{d_{k}}}\right) V
\end{align}
We use two sets of projections, $Q_s$, $K_s$, $V_s$ to compute attention scores of sliding window attention, and  $Q_g$, $K_g$, $V_g$ to  compute attention scores for the global attention.
The additional projections provide flexibility to model the different types of attention, which we show is critical for best performance on downstream tasks. 
$Q_g$, $K_g$, $V_g$ are all initialized with values that match
 $Q_s$, $K_s$, $V_s$. 

\subsection{Implementation}
\label{sec:tvm}
In regular transformers, attention scores are computed as in Eqn.~\ref{eq:qkv}. 
The expensive operation is the matrix multiplication $Q K^{T}$ because both
$Q$ and $K$ have $n$ (sequence length) projections. 
For \model, the dilated sliding window attention computes only a fixed number of the diagonals 
of $Q K^{T}$. As shown in Fig.~\ref{fig:tvm}, this results in a linear increase 
in memory usage compared to quadratic increase for full self-attention.
However, implementing it requires a form of banded matrix multiplication that is not supported in 
existing deep learning libraries like PyTorch/Tensorflow.
Fig.~\ref{fig:tvm} compares the performance of three different ways of implementing it:
\texttt{loop} is a memory efficient PyTorch implementation that supports dilation but is unusably slow and only used for testing;
\texttt{chunks} only supports the non-dilated case and is used for 
the pretraining/finetuning setting;
and \texttt{cuda} is our fully functioning highly optimized custom CUDA kernel implemented using TVM~\cite{tvm} and used for the language modeling experiments (see  Appendix~\ref{sec:tvm_details} for more details).

\section{Autoregressive Language Modeling}
Autoregressive or left-to-right language modeling is loosely defined as estimating the probability distribution of an existing token/character given its previous tokens/characters in an input sequence. 
This task is considered one of the fundamental tasks in natural language and recent prior work on modeling long sequences using transformers 
has relied on this task as their primary evaluation~\cite{transformerxl,compressive,adaptivespan}.
Similarly, we develop and evaluate our model on autoregressive language modeling. 

\subsection{Attention Pattern}
\label{sec:charlm_attn}
For autoregressive language modeling we use our dilated 
sliding window attention. 
Following~\citet{adaptivespan} we use differing window sizes across the layers. In particular, we use small window sizes for the lower layers and increase window sizes 
as we move to higher layers. This allows the top layers to learn higher-level representation of the entire sequence while having the lower layers capture local information. In addition, it provides balance between efficiency (smaller window sizes are less computationally expensive due to fewer nonzero values)
and performance (larger window sizes have richer representation power and often result in performance improvements). 

We do not use dilated sliding windows for lower layers to maximize their capacity to learn and utilize the immediate local context.
For the higher layers, we use a small amount of increasing dilation only on 2 heads.
This gives the model the ability to directly attend to distant tokens without sacrificing local context.

    

\subsection{Experiment Setup}

To compare to prior work we focus on character-level LM \cite[\texttt{text8} and \texttt{enwik8};][]{text8}.

\paragraph{Training}
Ideally, we would like to train our model on the largest window size and sequence length we can fit in a modern GPU memory. However, we found that 
the model needs a large number of gradient updates to learn the local context
first, before learning to utilize longer context. 
To accommodate this, we adopt a staged training procedure where we increase the attention window size and sequence length across multiple training phases. In particular, in the first phase we start with a short sequence length and window size, then on each subsequent phase, we double the window size and the sequence length, and halve the learning rate. 
This makes training fast, while keeping the slow part 
(longest sequences and window sizes) to the end. 
We train the model over 5 total phases with starting sequence length of 2,048 and ending sequence length of 23,040 
on the last phase (see Appendix~\ref{sec:charlmapp} for detailed configurations of each phase, and for all other hyperparameters).

\paragraph{Evaluation}
We evaluate with sequences of length 32,256.  Following \citet{transformerxl}, we split the dataset into overlapping sequences 
of size 32,256 with a step of size 512, and report the performance on the last 512 tokens on the sequence.

\subsubsection{Results}
Tab.~\ref{tab:charlm-small} and~\ref{tab:charlm-large} summarize evaluation results on \texttt{text8} and \texttt{enwik8} datasets. 
We achieve a new state-of-the-art on both \texttt{text8} and \texttt{enwik8} using the small models with BPC of \textbf{1.10} and \textbf{1.00} on \texttt{text8} and \texttt{enwik8} respectively, demonstrating the effectiveness of our model.

For large models, given how expensive these experiments are, 
and following recent work~\cite{reformer,compressive}, 
we are only evaluating on \texttt{enwik8}. 
Tab.~\ref{tab:charlm-large} shows that \model
outperforms the comparable Transformer-XL model, 
matches the performance of the comparable Sparse Transformer~\cite{sparseOpenai},
and matches or slightly underperforms recent models 
that have more than twice the number of parameters.
It is worth noting that Adaptive Span~\cite{adaptivespan}
and Compressive Transformer~\cite{compressive}
are not good fit for the pretraining-finetuning paradigm as discussed in 
\S\ref{sec:related}.

\begin{table}[t]
    \centering
    \small
    \begin{tabular}{@{}lrrr@{}}
    \toprule
    Model & \#Param & Dev &  Test \\
    \midrule
    \textbf{Dataset} \texttt{text8}& &  & \\
    T12~\cite{AlRfou2018CharacterLevelLM} & 44M &  -    & 1.18 \\
    Adaptive \cite{adaptivespan}                        & 38M & 1.05  & 1.11 \\
    BP-Transformer \cite{BPTransformer}                        & 39M &    -  & 1.11 \\
    Our \model                            & 41M & 1.04	& \textbf{1.10} \\
    [0.6ex]
\hdashline[0.2pt/0.2pt]\noalign{\vskip 0.6ex}
    \textbf{Dataset} \texttt{enwik8}&  & \\
    T12~\cite{AlRfou2018CharacterLevelLM} & 44M &  -    & 1.11 \\
    Transformer-XL \cite{transformerxl}                       & 41M &  -    & 1.06 \\
    Reformer \cite{reformer}                              &  -  &  -    & 1.05 \\
    Adaptive \cite{adaptivespan}                         & 39M & 1.04  & 1.02 \\
    BP-Transformer \cite{BPTransformer}                        & 38M &    -  & 1.02 \\
    Our \model                            & 41M & 1.02	& \textbf{1.00} \\
    \bottomrule
    \end{tabular}
    \caption{\emph{Small} model BPC on \texttt{text8} \& \texttt{enwik8}}
    \label{tab:charlm-small}
\end{table}

\begin{table}[t]
    \centering
    \small
    \begin{tabular}{@{}lrrr@{}}
    \toprule
    Model & \#Param & Test BPC \\
    \midrule
    Transformer-XL (18 layers)           & 88M  &  1.03 \\
    Sparse~\cite{sparseOpenai}                  &$\approx$100M &  0.99 \\
    Transformer-XL (24 layers)           & 277M &  0.99 \\
    Adaptive~\cite{adaptivespan}                        & 209M &  0.98 \\
    Compressive~\cite{compressive}              & 277M &  0.97 \\
    Routing~\cite{roy2020efficient} & $\approx$223M & 0.99 \\
    Our \model                           & 102M &  0.99 \\
    \bottomrule
    \end{tabular}
    \caption{Performance of \emph{large} models on \texttt{enwik8}}
    \label{tab:charlm-large}
\end{table}

\subsubsection{Ablation Study}

\begin{table}[t]
    \centering
    \small
    \begin{tabular}{@{}lr@{}}
    \toprule
    Model & Dev BPC \\
    \midrule
    Decreasing $w$ (from 512 to 32) & 1.24 \\
    Fixed $w$ (= 230) &  1.23 \\
    Increasing $w$ (from 32 to 512) &\textbf{ 1.21} \\
    \midrule
    No Dilation &  1.21 \\
    Dilation on 2 heads &  \textbf{1.20} \\
    \bottomrule
    \end{tabular}
    \caption{Top: changing window size
    across layers. Bottom: with/without dilation (@ 150K steps on phase1)}
    \label{tab:ablation_charlm}
\end{table}



To show the importance of the design choices of our attention patterns, we tried different variants and report their controlled experiment results. 
To make the ablation study more manageable, we train each configuration for 150K steps\footnote{
One caveat is that the ordering of end performance will not agree with that at step 150K.
However, this approximation saves the huge cost of running every experiment to completion.}
with phase 1 configuration on a small model on \texttt{text8}, then report the BPC performance on the dev set. 

The top of Tab.~\ref{tab:ablation_charlm} demonstrates the impact of different ways of configuring the window sizes per layer. We observe that increasing the window size from the bottom 
to the top layer leads to the best performance, arranging them in the reverse way leads to worse performance,
and using a fixed window size (the average of window sizes of the other configuration) leads to a performance that it is in between.
The bottom of Tab.~\ref{tab:ablation_charlm} shows the impact of adding dilation.
Adding some dilation to two heads leads to some improvement compared with 
no dilation at all. 



\section{Pretraining and Finetuning}
\label{sec:pretrain}

Current state-of-the-art systems for many NLP tasks finetune a pretrained model with task supervision (e.g. BERT).
One of our main motivations is to develop such a model suitable for long document tasks.
To do so, we pretrained \model on a document corpus and finetune it for six tasks, including classification, QA and coreference resolution.
The resulting model can process sequences up to 4,096 tokens long (8 times longer than BERT)\footnote{Sequences up to 16K are possible on current GPUs.}.



We pretrain \model with masked language modeling (MLM), where the goal is to recover randomly masked tokens in a sequence. 
Since MLM pretraining is expensive, we continue pretraining from the RoBERTa~\cite{roberta} released checkpoint, while only making the minimal changes necessary to support \model's attention mechanism.
Note that our attention pattern can be plugged into any pretrained transformer model without the need to change the model architecture. 


\paragraph{Attention Pattern}
We use sliding window attention with window size of 512, therefore using the same amount of computation as RoBERTa.\footnote{Adding dilation on a few heads as in \S\ref{sec:charlm_attn} hurt performance, likely because it is not compatible with the pretrained RoBERTa weights. Retraining such model from scratch might be needed to improve performance.
}

\paragraph{Position Embeddings}
RoBERTa uses learned absolute position embeddings with the maximum 
position being 512. To support longer documents, we add extra position embeddings to support up to position 4,096. 
To leverage RoBERTa's pretrained weights, instead of randomly initializing the new position embeddings, we initialize them by copying the 512 position
embeddings from RoBERTa multiple times as analysis of BERT's attention heads shows a strong learned bias to attending to local context, including the previous or next token \cite{Clark2019WhatDB}.  Using the copy initialization preserves this local structure everywhere except at the partition boundaries.
Despite its simplicity, we found this to be a very effective (see Tab.~\ref{tab:roberta}), allowing \model pretraining to rapidly converge with a small number of gradient updates.

\begin{table}[t]
    \centering
    \small
    \begin{tabular}{@{}lrr@{}}
    \toprule
    Model &  base & large\\
    \midrule
    RoBERTa (seqlen: 512)  & 1.846 & 1.496 \\
    \model  (seqlen: \seqlen) & 10.299 & 8.738 \\
    ~~~ + copy position embeddings & 1.957 & 1.597\\
    ~~~~~~ + 2K gradient updates & 1.753 & 1.414 \\
    ~~~~~~ + 65K gradient updates & 1.705 & 1.358\\
    \model (train extra pos. embed. only) & 1.850 & 1.504 \\
    \bottomrule
    \end{tabular}
    \caption{MLM BPC for RoBERTa and various pretrained \model configurations. 
    }
    \label{tab:roberta}
\end{table}

\paragraph{Continued MLM Pretraining}

We pretrain \model using fairseq \cite{ott2019fairseq} on a corpus of long documents that we compiled (see Appendix~\ref{sec:mlm_data} for corpus details).
We train two model sizes, a base model and a large model.
Both models are trained for 65K gradient updates with sequences length \seqlen, batch size 64 ($2^{18}$ tokens), maximum learning rate of 3e-5, linear warmup of 500 steps, followed by a power 3 polynomial decay. The rest of the hyperparameters are the same as RoBERTa. 

\begin{table}[t]
    \centering
    \small
    \setlength{\tabcolsep}{3pt} 
    \begin{tabular}{@{}lrrrrrr@{}}
    \toprule
    Wordpieces & WH & TQA & HQA & ON & IMDB & HY \\ \midrule
    avg.    & 1,535 & 6,589 & 1,316 & 506 & 300 & 705   \\
    95th pctl.     & 3,627 & 17,126 & 1,889 & 1,147 & 705 & 1,975 \\ \bottomrule
    \end{tabular}
    \caption{Average and 95th percentile of context length of datasets in wordpieces. WH: WikiHop, TQA: TriviaQA, HQA: HotpotQA, ON: OntoNotes, HY: Hyperpartisan news} 
    \label{tab:doc-len}
\end{table}

\begin{table*}[t]
    \centering
    \small
    \begin{tabular}{@{}lrrrrrr@{}}
    \toprule
     & \multicolumn{3}{c}{QA} & \multicolumn{1}{c}{Coref.} & \multicolumn{2}{c}{Classification} \\
     \cmidrule(lr){2-4} \cmidrule(lr){5-5} \cmidrule(l){6-7}
    Model &  WikiHop & TriviaQA & HotpotQA & OntoNotes & IMDB & Hyperpartisan \\
    \midrule
    RoBERTa-base    & 72.4 & 74.3 & 63.5  & 78.4 & 95.3 & 87.4 \\
    \model-base     & \textbf{75.0} & \textbf{75.2} & \textbf{64.4} & \textbf{78.6} & \textbf{95.7} & \textbf{94.8} \\
    \bottomrule
    \end{tabular}
    \caption{Summary of finetuning results on QA,  
    coreference resolution, and document classification. 
    Results are on the development sets comparing our
    \model-base with RoBERTa-base.
    TriviaQA, Hyperpartisan metrics are F1, WikiHop and IMDB use accuracy, HotpotQA is joint F1, OntoNotes is average F1.
    }
    \label{tab:finetune}
\end{table*}

Tab.~\ref{tab:roberta} shows the BPC on the development 
set of our training corpus. The first row shows a 1.846 BPC using 
RoBERTa-base, which is comparable to the 1.880 BPC reported
on the RoBERTa paper on their corpus. This indicates 
our training corpus is from a distribution close to that used to train RoBERTa.
The following two rows show the performance of \model
before pretraining 
with randomly initialized position embeddings and with 
copied position embeddings. The significant difference indicates the importance of the copy initialization, and the relative small difference between the RoBERTa BPC and the initialized BPC indicates that our sliding window attention is working well with the RoBERTa weights.
The following two rows show the impact of continuing pretraining. 
Traininig for 2K steps improves BPC from 1.957 to 1.753, which further decreases to 1.705 after 65K steps, demonstrating the model is learning to better utilize the sliding window attention and longer context.
Similar patterns are observed with RoBERTa-large and \model-large.

\paragraph{Frozen RoBERTa Weights}
We also pretrained \model while freezing all RoBERTa weights, and only training the new position embeddings.  The motivation for this configuration is to perfectly preserve 
the RoBERTa performance on short documents.
This configuration has a BPC of 1.850 (down from 1.957 at initialization), but higher than 1.705 where all the weights 
are trainable.

\section{Tasks}

We apply \model to multiple long document tasks, including QA, coreference resolution and classification.  Tab.~\ref{tab:doc-len} shows the evaluation datasets have contexts significantly longer than 512 wordpieces.
Our primary goal is to evaluate whether our attention mechanism can act as a replacement for the standard self-attention mechanism in BERT style models, and to perform controlled trials against a strong baseline.
We are also interested in evaluating whether we can replace complicated task specific models necessitated by BERT's limited context with simpler models that just concatenate all available context into a single sequence.

Our baseline is a RoBERTa based model that breaks the context into the longest possible segment, passes each individually through RoBERTa, and concatenates the activations for further processing.
For QA tasks, we also concatenate the question to each segment so that RoBERTa can condition it's contextual representations of the context on the question.
The \model variant replaces the RoBERTa self-attention mechanism with our windowed attention used during pretraining, plus a task motivated global attention.  The global attention uses additional linear projections (\S\ref{sec:attn}).


\subsection{Question answering}
We used three datasets: WikiHop \cite{Welbl2018ConstructingDF-Wikihop}, TriviaQA \cite[][Wikipedia setting]{Joshi2017TriviaQAAL}, and HotpotQA, \cite[][distractor setting]{Yang2018-HotpotQAAD}.\footnote{We use the full version of TriviaQA and HotpotQA, not the simplified versions in MRQA~\cite{mrqa}.} 


For WikiHop and TriviaQA we follow the simple QA model of BERT~\cite{bert}, and concatenate question and documents into one long sequence, run it through \model, then have a dataset-specific prediction layer. WikiHop uses a classification layer for the candidate while TriviaQA uses the loss function of~\citet{Clark2017SimpleAE} to predict answer span.  We include global attention to question tokens and answer candidates for WikiHop and to question tokens for TriviaQA.

HotpotQA is a multihop QA dataset that involves extracting answer spans and evidence sentences from 10 Wikipedia paragraphs, 2 of which are relevant and the rest are distractors. We use a two-stage model
that first selects the most relevant paragraphs then passes them to a second stage for answer extraction. Both stages concatenate question and context into one sequence, run it through \model, then use task-specific prediction layers. 
We train the models in a multi-task way to predict relevant paragraphs, evidence sentences, answer spans and question types (yes/no/span) jointly.
Note that this model is simpler than recent SOTA models that include complex task-specific architectures (e.g.,~\cite{Tu2019SelectAA,Chen2019MultihopQA,Tu2020GraphSN,quark2020}). 
See Appendix~\ref{sec:taskdetails} for further details about the models and hyperparameters.

\subsection{Coreference Resolution}
We use OntoNotes \cite{pradhan-etal-2012-conll}, and the model from \citet{joshi-etal-2019-bert}, a modification of the system from \citet{lee-etal-2018-higher} to replace ELMo with BERT.
The \model system is a straightforward adaption of the baseline model by replacing RoBERTa with \model and extending the sequence length.
We didn't use global attention for this task. 

\subsection{Document Classification }
We evaluate on IMDB \cite{imdb} and Hyperpartisan news detection \cite{hyperpartisan} datasets.\footnote{
For Hyperpartisan we split the training data into 80/10/10 train/dev/test sets, and report mean F1 across five seeds.} IMDB is a standard sentiment classification datasets consisting of movie reviews. While most documents in this dataset are short, about 13.6\% of them are larger than 512 wordpieces (Tab.~\ref{tab:doc-len}).
Documents in Hyperpartisan are relatively long, and it is small with only 645 documents making it a good test for \model's ability to adapt to limited data. We use global attention 
on the \texttt{[CLS]} token. 

\subsection{Results}

\paragraph{Main Result}
Tab.~\ref{tab:finetune} summarizes the results of all our finetuning experiments. We observe that \model consistently outperforms the RoBERTa baseline. 
Its performance gain is especially obvious for tasks that require long context such as WikiHop and Hyperpartisan.
For TriviaQA, the improvement is more modest as the local context is often sufficient to answer the question.
In the case of HotpotQA, the supporting fact auxiliary supervision allows models to easily find relevant contexts and then focus on local context, leading to smaller gains.  This is contrasted with WikiHop that only includes distant supervision of intermediate reasoning chains, where our approach excels by reasoning over the entire context.
On the IMDB and OntoNotes datasets the performance gains are smaller. For IMDB, the majority of the dataset consists of short documents and thus it is expected to see smaller improvements.
For OntoNotes, we found that the distance between any two mentions is typically quite small so that a baseline that processes smaller chunks separately is able to stitch together mentions into coreference chains without considering cross chunk interactions.



\paragraph{\model-large for QA}
We also evaluate the performance of \model-large on long context QA tasks. Tab.~\ref{tab:leaderboard}
shows that our \model-large achieves new state-of-the-art
results\footnote{At submission time, May 2020. Later, BigBird \cite{Zaheer2020BigBT} improved leaderboard results on these datasets. There are confounding factors such as using 16X more compute in BigBird's pretraining compared with Longformer, potentially affecting the performance.} on WikiHop and TriviaQA by large margins (3.6 and 4 points respectively), and for HotpotQA, it underperforms the current state-of-the-art \cite{hotpotqasota} by a point.
Tab.~\ref{tab:hotpotqa} shows the detailed results of HotpotQA compared with published and unpublished concurrent models. \model places second on the published leaderboard, outperforming all other published results except for HGN~\cite{hotpotqasota}. All published top performing models in this task \cite{Tu2019SelectAA,hotpotqasota,Shao2020IsGS} use GNNs \cite{kipf2017semi} or graph network of entities, which seem to encode an important inductive bias for the task and can potentially improve our results further. 
Nevertheless, \model performs strongly outperforming all other methods including the recent non-GNN methods~\cite{Gla2019SpanSP,Shao2020IsGS,quark2020}.

\begin{table}[t]
\centering
\small
\begin{tabular}{@{}lrrr@{}}
\toprule
Model            & WikiHop & TriviaQA & HotpotQA \\ 
\midrule
Current$^*$ SOTA                   & 78.3 & 73.3 & \textbf{74.2} \\
\model-large    & \textbf{81.9} & \textbf{77.3} & 73.2\\
\bottomrule
\end{tabular}
\caption{Leaderboard results of \model-large at time of submission (May 2020). All numbers are F1 scores.
} 
\label{tab:leaderboard}
\end{table}

\begin{table}[t]
\centering
\small
\setlength{\tabcolsep}{4pt} 
\begin{tabular}{@{}lrrr@{}}
\toprule
Model            & ans. & supp. & joint \\ \midrule
TAP 2 (ensemble)~\cite{Gla2019SpanSP}     &  79.8	   & 86.7	    & 70.7  \\
SAE \cite{Tu2019SelectAA} & 79.6 & 86.7 & 71.4 \\
Quark (dev) \cite{quark2020} & 81.2 & 87.0 & 72.3  \\
C2F Reader~\cite{Shao2020IsGS} & 81.2	& 87.6 &	72.8 \\
[0.5ex]
\hdashline[0.4pt/2pt]\noalign{\vskip 0.5ex}
\model-large & 81.3 &	88.3 &	73.2 \\ 
[0.5ex]
\hdashline[0.4pt/2pt]\noalign{\vskip 0.5ex}
ETC-large$^\dag$ \cite{ainslie-etal-2020-etc} & 81.2 & 89.1 & 73.6 \\
GSAN-large$^\dag$ & 81.6 & 88.7 & 73.9 \\
HGN-large~\cite{hotpotqasota}       & 82.2   & 88.5	    & 74.2  \\
\bottomrule
\end{tabular}
\caption{
HotpotQA results in distractor setting test set. Quark's test results are not available. All numbers are F1 scores. $^\dag$ shows contemporaneous leaderboard submissions.
} 
\label{tab:hotpotqa}
\end{table}

\subsection{Ablations on WikiHop}

\begin{table}[t]
    \centering
    \small
    \setlength{\tabcolsep}{-2pt}
    \begin{tabular}{lr}
    \toprule
    Model  & Accuracy / $\Delta$ \\
    \midrule
    \model (seqlen: 4,096) & 73.8 \\
    [0.6ex]
    \hdashline[0.2pt/0.4pt]\noalign{\vskip 0.6ex}
    RoBERTa-base (seqlen: 512) & 72.4 / -1.4\\
    \model (seqlen: 4,096, 15 epochs) & 75.0 / +1.2\\
    \model (seqlen: 512, attention: $n^2$) & 71.7 / -2.1 \\
    \model (seqlen: 2,048) & 73.1 / -0.7 \\
    \model (no MLM pretraining) & 73.2 / -0.6 \\
    \model (no linear proj.) & 72.2 / -1.6 \\
    \model (no linear proj. no global atten.) & 65.5 / -8.3 \\
    \model (pretrain extra position embed. only) & 73.5 / -0.3 \\
    \bottomrule
    \end{tabular}
    \caption{WikiHop development set ablations
    }
    \label{tab:finetune_ablation}
\end{table}

Tab.~\ref{tab:finetune_ablation} presents an ablation study for WikiHop on the development set. All results use \model-base, fine-tuned for five epochs with identical hyperparameters except where noted.  \model benefits from longer sequences, global attention, separate projection matrices for global attention, MLM pretraining, and longer training.  In addition, when configured as in RoBERTa-base (seqlen: 512, and $n^2$ attention) \model performs slightly worse then RoBERTa-base, confirming that performance gains are not due to additional pretraining.  Performance drops slightly when using the RoBERTa model pretrained when only unfreezing the additional position embeddings, showing that \model can learn to use long range context in task specific fine-tuning with large training datasets such as WikiHop.



\section{Longformer-Encoder-Decoder (LED)}
\label{sec:led}

The original Transformer \cite{Vaswani2017AttentionIA} consisted of an encoder-decoder architecture, intended for
sequence-to-sequence tasks \cite{Sutskever2014SequenceTS}, such as summarization and translation. 
While encoder-only Transformers are effective on a variety of NLP tasks, pre-trained encoder-decoder Transformer models
(e.g.\ BART \cite{lewis-etal-2020-bart} and T5 \cite{Raffel2020ExploringTL})
have achieved strong results on tasks like summarization. Yet, such models can't efficiently scale to seq2seq tasks with longer inputs.

To facilitate modeling long sequences for seq2seq learning, we propose a Longformer variant that has both the encoder and decoder Transformer stacks but instead of the full self-attention in the encoder, it uses the efficient local+global attention pattern of the Longformer.  The decoder uses the full self-attention to the entire encoded tokens and to previously decoded locations. We call this model Longformer-Encoder-Decoder (LED) which scales linearly with the input.
Since pre-training LED is expensive, we initialize LED parameters from the BART, and follow BART's exact architecture in terms of number of layers and hidden sizes. The only difference is that to process longer inputs, we extend position embedding to 16K tokens (up from BART's 1K tokens) and we initialize the new position embedding matrix by repeatedly copying BART's 1K position embeddings 16 times as in Section~\ref{sec:pretrain} for RoBERTa. Following BART, we release two model sizes, LED-base and LED-large, which respectively have 6 and 12 layers in both encoder and decoder stacks.


We evaluate LED on the summarization task using the arXiv summarization dataset \cite{arxiv2018} which focuses on long document summarization in the scientific domain. The 90th percentile of document lengths is 14.5K tokens, making it an appropriate testbed for evaluating LED. LED's encoder reads the document and its decoder generates the output summary. The encoder uses local attention with window size 1,024 tokens and global attention on the first \texttt{<s>} token. The decoder uses full attention to the entire encoder and previously decoded locations. As standard in seq2seq models, LED is trained using teacher forcing on gold training summaries and uses beam search at inference.

\begin{table}[]
\centering
\small
\begin{tabular}{@{}lrrr@{}}
\toprule
                & R-1   & R-2   & R-L   \\ \midrule
Discourse-aware \citeyearpar{arxiv2018} & 35.80 & 11.05 & 31.80 \\
Extr-Abst-TLM \citeyearpar{Subramanian2020OnEA}   & 41.62 & 14.69 & 38.03 \\
Dancer  \citeyearpar{Gidiotis2020ADA}        & 42.70 & 16.54 & 38.44 \\
Pegasus \citeyearpar{zhang2019pegasus}        & 44.21 & 16.95 & 38.83 \\ 
LED-large (seqlen: 4,096) (ours) &   44.40      &  17.94     & 39.76 \\
BigBird (seqlen: 4,096) \citeyearpar{Zaheer2020BigBT}        & \bf{46.63} & 19.02 & 41.77 \\
\hdashline[.4pt/1pt]
LED-large (seqlen: 16,384) (ours) & \bf{46.63}      &  \textbf{19.62}     & \bf{41.83}       \\ \bottomrule
\end{tabular}
\caption{Summarization results of Longformer-Encoder-Decoder (LED) on the arXiv dataset.  Metrics from left to right are ROUGE-1, ROUGE-2 and ROUGE-L.}\label{tab:summarization}
\end{table}

\begin{figure}[t!]
    \centering
    \includegraphics[width=0.7\linewidth]{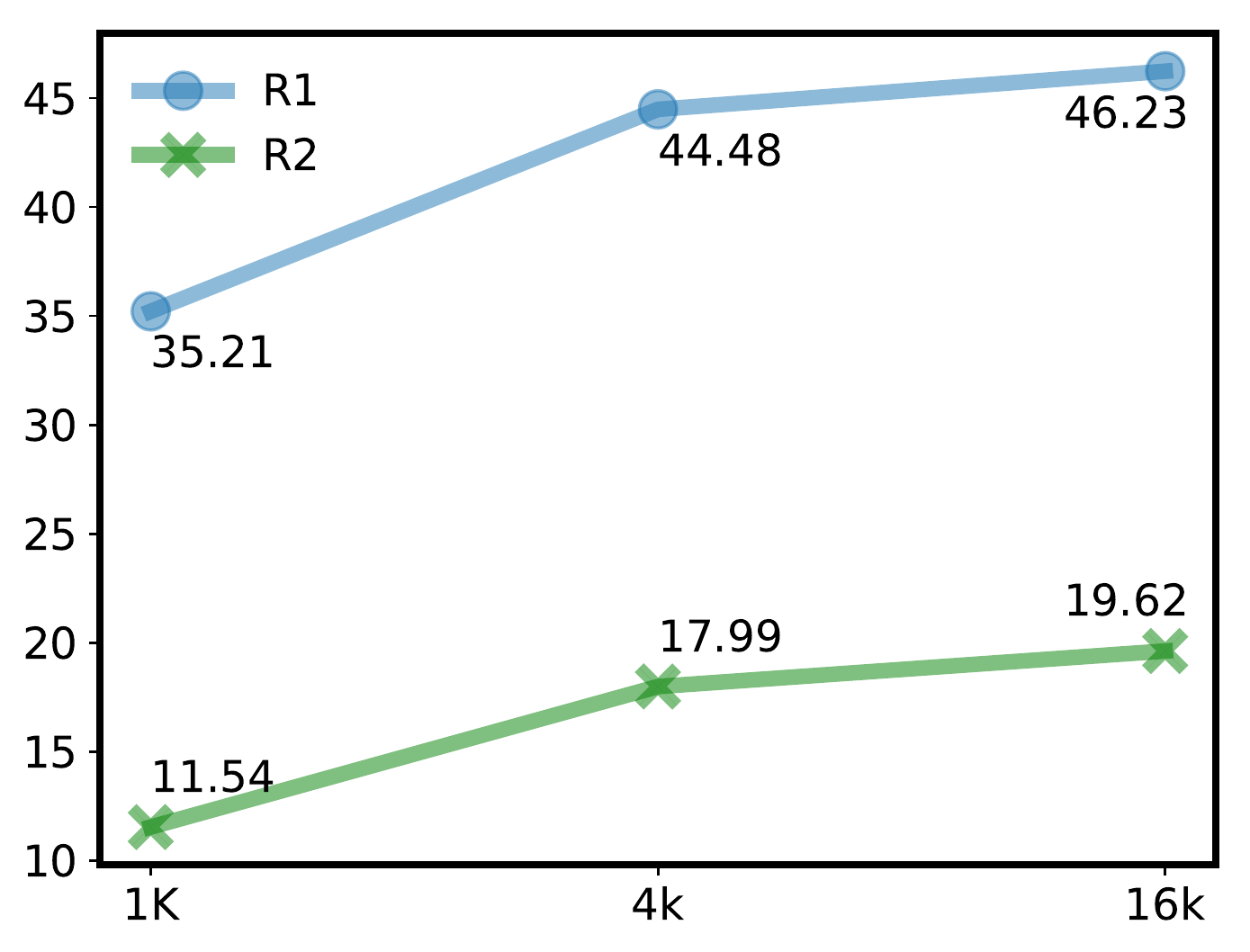}
    \caption{
    ROUGE-1 and ROUGE-2 of LED when varying the input size (arXiv validation set).
    }
    \label{fig:led}
\end{figure}

Tab. \ref{tab:summarization} demonstrates the results of LED-large 16K on the arXiv summarization task. This model is merely initialized from BART, with no additional pre-training. We observe that LED achieves state-of-the-art results on arXiv, slightly outperforming BigBird \cite{Zaheer2020BigBT}.
Note that the BigBird summarization model supports sequence length of 4K tokens but starts from and continues pre-training Pegasus \cite{zhang2019pegasus}, a model specifically designed and pre-trained for summarization.
With no pre-training or task-specific initialization, but with ability to process longer inputs, LED can slightly outperform BigBird. Further improvements should be possible through pre-training of LED.
Fig. \ref{fig:led} further illustrates the importance of sequence length showing the ablility to process longer input significantly improves the results.








\section{Conclusion and Future Work}
We present \model, a transformer-based model that is scalable for processing long documents and that makes it easy to perform a wide 
range of document-level NLP tasks without chunking/shortening the long input and without complex architecture to combine information across these chunks. 
\model employs an attention pattern that combines local and global
information while also scaling linearly with the sequence length.
\model achieves state-of-the-art results on the character-level
language modeling tasks of \texttt{text8} and \texttt{enwik8}. 
When pretrained, 
\model consistently outperforms RoBERTa on long document tasks and
sets new state-of-the-art results on WikiHop and TriviaQA. 
We further present LED, an encoder-decoder variant of Longformer for modeling sequence-to-sequence tasks, and achieve state-of-the-art results on the arXiv long document summarization task. 
For future work, we would like to study other pretraining objectives, especially for LED, increase the sequence length, and explore other tasks that might benefit from our model. 


\section*{Acknowledgment}
We would like to thank Noah Smith, Dan Weld, Dirk Groeneveld, Kyle Lo, Daniel King and Doug Downey for helpful discussions and feedback, and the AI2 infrastructure team for technical support. 

\bibliography{emnlp2020}
\bibliographystyle{acl_natbib}
\clearpage

\appendix
\section{Implementation Details}
\label{sec:tvm_details}

Implementing \model's dilated sliding window attention requires a form
of banded matrix multiplication (matrix multiplication where the output
is all zero except certain diagonals) that is not directly supported in 
existing deep learning libraries like PyTorch/Tensorflow. 
Fig.~\ref{fig:tvm} compares the runtime and memory of three different ways of implementing
it. \\
\texttt{\model-loop} is a naive implementation that computes each diagonal separately in a loop. 
It is memory efficient because it only computes the non-zero values, but 
it is unusably slow. We only use it for testing because it is easy to implement but don't use 
it to run experiments. \\
\texttt{\model-chunks} only supports the non-dilated case. It chunks $Q$ and $K$ into 
overlapping blocks of size $w$ and overlap of size $\frac{1}{2}w$, multiplies the blocks, 
then mask out the diagonals. This is very compute efficient because it uses
a single matrix multiplication operation from PyTorch, but it consumes 2x the amount of memory
a perfectly optimized implementation should consume because it computes some of the zero values.
Because of the compute efficiency, this implementation is most suitable for the 
pretrain/finetune case. We didn't find the increase in memory to be a problem for this setting. \\
\texttt{\model-cuda} is a custom CUDA kernel that we implement using 
TVM~\cite{tvm}. It is a fully functioning implementation of our attention (not limited as \texttt{\model-chunks}), it is the most memory efficient, 
and it is as fast as the highly optimized full self-attention.\footnote{It is worth noting that theoretically, a perfectly optimized \texttt{\model-cuda} should be faster than the $n^2$ computation. 
However, achieving this level of performance requires special knowledge of low-level GPU programming, similar to implementing a highly optimized matrix multiplication. Our current implementation is sufficiently fast and practical to use.} We mainly use this implementation for the 
autoregressive language modeling experiments because of the memory efficiency (allows the longest 
sequences) and the support of dilation (needed for character-LM experiments).

\paragraph{Tensor Virtual Machine (TVM)}
We build our custom CUDA kernel using TVM~\cite{tvm}, a deep learning compiler stack that compiles
high level description of a function into optimized device-specific code.
Using TVM, we describe our banded matrix multiplication in 
high-level python constructs, then TVM generates the corresponding CUDA code
and compiles it for GPUs.

\section{Character LM Hyperparameters}
\label{sec:charlmapp}

We evaluate on \texttt{text8} and \texttt{enwik8}, both contain 100M 
characters from Wikipedia split into 90M, 5M, 5M for train, dev, test. 
Our model only specifies how the self-attention component works, and it is 
agnostic to the other design choices for the transformer model. 
Our implementation is based on the Transformer-XL~\cite{transformerxl}
code\footnote{\url{https://github.com/kimiyoung/transformer-xl}} 
with the memory mechanism disabled. 
We use relative position embeddings with sinusoidal weights as in~\citet{transformerxl}. 
We use two different model sizes, a small (12 layers, 512 hidden size) model as in~\citet{transformerxl}, 
and a large (30 layers, 512 hidden size) model as in~\citet{sparseOpenai}.
We employed mixed precision training (floating points 16 and 32) using apex\footnote{\url{https://github.com/NVIDIA/apex}} to reduce memory consumption and speed-up training. However,  we kept the attention computation in fp32
to avoid numerical instability issues.\footnote{We found that using fp16 in attention operation results in floating point overflow and NaNs in later stages of training.}
We used gradient checkpointing~\cite{gradckpt} to reduce memory usage, and ran our experiments on 48GB RTX8000 GPUs. 
All hyperparameters and stage configurations are listed in Tab.~\ref{tab:char-hyperparams}. 
Our CUDA kernel supports the autoregressive mode where each token 
attends to a window of previous tokens only. Our implementation 
also includes a version of the relative position embedding
that is compatible with our dilated sliding window attention.

We ran the small model experiments on 4 RTX8000 GPUs for 16 days.
For the large model, we ran experiments on 8 RTX8000 GPUs for 
13 days. 
Most of our hyperparameter search is similar to the ablation in Tab.~\ref{tab:ablation_charlm} where we run the configuration for 150K steps
on \texttt{text8}.
We experimented with absolute position embeddings and learned position embeddings,
dropout values of [0.1, 0.2] (small model) and [0.1, 0.4] (large model), 
pre-layernorm and post-layernorm~\cite{layernorm}, learning rate (LR) of phase1 of values 
[2.5e-5, 5e-4, 1e-4]
constant and cosine LR schedules, 
and different configurations for dilation (on all heads, on 2 heads, no dilation).
Number of gradient updates/phase
reported in Tab.~\ref{tab:char-hyperparams} is determined by running each phase until the validation BPC stops getting better.

\begin{table*}
 \centering
 \small
\begin{tabular}{@{}ll@{}}
\toprule
Param & Value \\
\midrule
Position Embeddings & Relative and Sinusoidal as in~\citet{transformerxl} \\
Small model config & 12 layers, 8 heads, 512 hidden size as in~\citet{transformerxl} \\
Large model config & 30 layers, 8 heads, 512 hidden size as in~\citet{sparseOpenai} \\
Optimizer & AdamW \\
Dropout & 0.2 (small model), 0.4  (large model) \\
Gradient clipping & 0.25 \\
Weight Decay & 0.01 \\
Layernorm Location & pre-layernorm~\cite{layernorm} \\
Activation & GeLU \\
Number of phases & 5 \\
Phase 1 window sizes & 32 (bottom layer) - 8,192 (top layer) \\
Phase 5 window sizes & 512 (bottom layer) -  (top layer) \\
Phase 1 sequence length & 2,048 \\
Phase 5 sequence length & 23,040  (gpu memory limit) \\
Phase 1 LR & 0.00025 \\
Phase 5 LR & 000015625 \\
Batch size per phase & 32, 32, 16, 16, 16 \\
\#Steps per phase (small) & 430K, 50k, 50k, 35k, 5k \\
\#Steps per phase (large) & 350K, 25k, 10k, 5k, 5k \\
Warmup & 10\% of the phase steps with maximum 10K steps \\
LR scheduler & constant throughout each phase \\
Dilation (small model) & 0 (layers 0-5), 1 (layers 6-7), 2 (layers 8-9), 3 (layers 10-11) \\
Dilation (large model) & 0 (layers 0-14), 1 (layers 15-19), 2 (layers 20-24), 3 (layers 25-29) \\
Dilation heads & 2 heads only \\
\bottomrule
\end{tabular}
\caption{Hyperparameters for the best performing model for character-level language modeling}
\label{tab:char-hyperparams}
 \end{table*}

\section{Pretraining Data}
\label{sec:mlm_data}

\begin{table}[t]
\centering
\small
\begin{tabular}{@{}lrr@{}}
\toprule
Source    & Tokens & Avg doc len \\ \midrule
Books \cite{Zhu2015AligningBA}    & 0.5B              & 95.9K\\
English Wikipedia & 2.1B              & 506 \\
Realnews \cite{Zellers2019DefendingAN}  & 1.8B              & 1.7K\\
Stories \cite{Trinh2018ASM}  & 2.1B              & 7.8K \\ \bottomrule
\end{tabular}
\caption{Pretraining data}\label{tab:pretrain-data}
\end{table}

In order to allow the model to learn long dependencies in pretraining, we compiled a corpus of long documents. Some of these data sources were also included in the original RoBERTa pretraining including the Books corpus \cite{Zhu2015AligningBA} plus English Wikipedia. We additionally included one third of a subset of the Realnews dataset \cite{Zellers2019DefendingAN} with documents longer than 1,200 tokens as well as one third of the Stories \cite{Trinh2018ASM} corpus. Our goal was to include a mix of long and short documents to both allow the model to learn longer dependencies while not to forget information from the original RoBERTa pretraining. The statistics of the pretraining data is shown in Tab.~\ref{tab:pretrain-data}.

\section{Task specific model details}
All the QA and classification models are implemented using PyTorch-Lightning\footnote{\url{https://github.com/PyTorchLightning/pytorch-lightning}}.  We use the official train/dev/test splits of all datasets except for the Hyperpartisan news which we randomely split into 80/10/10 for train/dev/test.

\label{sec:taskdetails}

\paragraph{WikiHop} Instances in WikiHop consist of: a question, answer candidates (ranging from two candidates to 79 candidates), supporting contexts (ranging from three paragraphs to 63 paragraphs), and the correct answer.  The dataset does not provide any intermediate annotation for the multihop reasoning chains, requiring models to instead infer them from the indirect answer supervision.

To prepare the data for input to \model and RoBERTa, we first tokenize the question, answer candidates, and support contexts using RoBERTa's wordpiece tokenizer.  Then we concatenate the question and answer candidates with special tokens as \texttt{[q] question [/q] [ent] candidate1 [/ent] ... [ent] candidateN [/ent]}.  The contexts are also concatenated using RoBERTa's document delimiter tokens as separators: \texttt{</s> context1 </s> ... </s> contextM </s>}.  The special tokens \texttt{[q], [/q], [ent], [/ent]} were added to the RoBERTa vocabulary and randomly initialized before task finetuning.

After preparing the input data, we compute activations from the top layer of each model as follows.
We take the question and answer candidates and concatenate them to as much context as possible up to the model sequence length (512 for RoBERTa, 4,096 for \model), run the sequence through the model, collect the output activations, and repeat until all of the context is exhausted (for all models except \model-large, where we just include the first 4,096 length sequence due to memory requirements).  Then all activations for all chunks are concatenated into one long sequence.  In the case of \model, we use global attention to the entire question and answer candidate sequence.

For prediction, we attach a linear layer to each \texttt{[ent]} that outputs a single logit, average over all logits for each candidate across the chunks, apply a softmax and use the cross entropy loss with the correct answer candidate.

Training used the Adam optimizer with linear warmup over 200 gradient updates to a maximum LR, and linear decay over the remainder of training.  We used gradient accumulation to effective batch size of 32 instances, checking the development accuracy every 250 gradient updates and reported the maximum development accuracy.  Other hyperparameters (dropout, weight decay) were identical to RoBERTa pretraining.

In general, we ran minimal hyperparameter trials, but for fair comparison between \model and RoBERTa ran an identical hyperparameter search with \model-base and RoBERTa-base.  This consisted of a grid search of LR in [2e-5, 3e-5, 5e-5] and number epochs in [5, 10, 15].  The best \model-base configuration used lr=3e-5, 15 epochs.  We ran two hyperparameter trials for \model-large, lr=3e-5 and number epochs in [5, 15] (the 5 epoch model had higher dev accuracy of 77.6, and was the single model submitted to the public leaderboard for test set evaluation).  All models were trained on a single RTX8000 GPU, with \model-base taking about a day for 5 epochs.

\paragraph{TriviaQA}
TriviaQA has more than 100K question, answer, document triplets for training. 
Documents are Wikipedia articles, and answers are named entities 
mentioned in the article. The span that answers the question is not annotated, 
but it is found using simple text matching. 

Similar to WikiHop, we tokenize the question and the document 
using RoBERTa's tokenizer, then form the input as \texttt{[s] question [/s] 
document [/s]}. We truncate the document at 4,096 wordpiece to avoid 
it being very slow. Afterwards, we get the activations from RoBERTa 
and Longformer similar to WikiHop (discussed above). 
We use global attention on all question tokens. 

For prediction, we add one layer that predicts the beginning and end of 
the answer span. Because of the distant supervision nature of the training data (no gold answer spans), we use the loss function of~\citet{Clark2017SimpleAE}
which works like an OR that the model only needs to get 
one answer span right, not all of them. 

Hyperparameters of the best configuration are listed in Tab.~\ref{tab:qa-hyperparams}. All other hyperparameters are similar to RoBERTa's. For hyperparameter search, we only tuned LR for the RoBERTa
baseline and tried rates [3e-5, 5e-5, 1e-4], then used the best, which is 3e-5, 
for all subsequent experiments with no further tuning. 
We trained the \model-large with the best configuration once and submitted its 
output to the leaderboard. 
We ran our experiments on 32GB V100 GPUs.
Small model takes 1 day to train on 4 GPUs, while large model takes 
1 day on 8 GPUs.

\paragraph{HotpotQA}
HotpotQA dataset involves answering questions from a set of 10 paragraphs from 10 different Wikipedia articles where 2 paragraphs are relevant to the question and the rest are distractors. It includes 2 tasks of answer span extraction and evidence sentence identification.
Our model for HotpotQA combines both answer span extraction and evidence extraction in one joint model. 
We found a higher performance using a two-stage \model model with similar setup that first identifies relevant paragraphs and then does find the final answer span and evidence.\footnote{The final dev performance of the two stage model improves over a single stage model by about 4.2 points on joint-F1 metric}  This is largely because removing the distracting paragraphs first reduces the noise for the final evidence and span detection as also found to be important by recent state-of-the-art methods in this dataset \cite{hotpotqasota}.
Similar to Wikihop and TriviaQA, to prepare the data for input to \model, we concatenate question and then all the 10 paragraphs in one long context. We particularly use the following input format with special tokens: ``\texttt{[CLS] [q] question [/q] $\langle$t$\rangle$ $\texttt{title}_{\texttt{1}}$ $\langle$/t$\rangle$} $\texttt{sent}_{\texttt{1,1}}$ \texttt{[s]} $\texttt{sent}_{\texttt{1,2}}$ \texttt{[s]} \texttt{...} \texttt{$\langle$t$\rangle$ $\texttt{title}_{\texttt{2}}$ $\langle$/t$\rangle$ }  $\texttt{sent}_{\texttt{2,1}}$ \texttt{[s]} $\texttt{sent}_{\texttt{2,2}}$ \texttt{[s]} \texttt{...}'' where \texttt{[q]}, \texttt{[/q]}, $\langle$\texttt{t}$\rangle$, $\langle$\texttt{/t}$\rangle$, \texttt{[s]}, \texttt{[p]} are special tokens representing, question start and end, paragraph title start and end, and sentence, respectively. The special tokens were added to the \model vocabulary and randomly initialized before task finetuning. For \model, we use global attention to question tokens, paragraph title start tokens as well as sentence tokens. The model includes additional feedforward layers on top of paragraph title start tokens for prediction of relevant paragraphs, as well as sentence tokens for predicting evidence sentences. After training the first stage model, we predict relevant paragraph scores for both training and development set. We then keep up to 5 paragraphs whose raw score is higher than a pre-specified threshold (-3.0), and remove the other paragraphs from the context. We then train the second stage model on the resulting shortened context. For answer span extraction we use BERT's QA model \cite{bert} with addition of a question type (yes/no/span) classification head over the first special token (\texttt{[CLS]}). For evidence extraction we apply 2 layer feedforward networks on top of the representations corresponding to sentence and paragraph tokens to get the corresponding evidence prediction scores and use binary cross entropy loss to train the model. At inference time for evidence extraction, we use a constrained decoding strategy similar to \citet{quark2020} that ensures that the evidence sentences come from exactly two paragraphs which is the setup of this dataset. We combine span, question classification, sentence, and paragraphs losses and train the model in a multitask way using linear combination of losses. Our experiments are done on RTX8000 GPUs and training each epoch takes approximately half a day on 4 GPUs.
We trained the model using Adam optimizer with linear warmup (1000 steps) and linear decay. We used minimal hyperparameter tuning using LRs of 3e-5 and 5e-5 and epochs of 3 to 7 and found the model with LR of 3e-5 and 5 epochs to work best. We conduct the same hyperparameter search for the RoBERTa baseline as well. The rest of hyperparameters are reported in Tab \ref{tab:qa-hyperparams}. 

\begin{table}[!h]
 \centering
 \small
\begin{tabular}{@{}lrrr@{}}
\toprule
Param & WikiHop & TriviaQA & HotpotQA \\
\midrule
Epochs & 15 & 5 & 5 \\
LR & 3e-5 & 3e-5 & 5e-5\\
Warmup steps & 200 & 1000 & 1000 \\
Batch size & 32 & 32 & 32 \\
Optimizer & Adam & Adam & Adam \\
\bottomrule
\end{tabular}
\caption{Hyperparameters of the QA models. All models use a similar scheduler with linear warmup and decay.}
\label{tab:qa-hyperparams}
 \end{table}

\paragraph{Coreference model details}
The coreference model is a straightforward adaptation of the coarse-to-fine BERT based model from \citet{joshi-etal-2019-bert}.  After preprocessing each document with the RoBERTa wordpiece tokenizer, it splits each document into non-overlapping segments up to the maximum sequence length, then concatenates the activations for the coarse-to-fine clustering stage that forms coreference clusters.
The maximum sequence length was 384 for RoBERTa-base, chosen after three trials from [256, 384, 512] using the default hyperparameters in the original  implementation.\footnote{https://github.com/mandarjoshi90/coref}  For \model-base the sequence length was 4,096.  Similar to the original implementation, different learning rates were used for the pretrained RoBERTa parameters and the randomly initialized task parameters.  Using a larger learning rate in the task parameters allows the optimizer to adjust them farther from their randomly initialized values without destroying the information in the pretrained RoBERTa parameters.  

Hyperparameter searches were minimal and consisted of grid searches of RoBERTa LR in [1e-5, 2e-5, 3e-5] and task LR in [1e-4, 2e-4, 3e-4] for both RoBERTa and \model for a fair comparison. The best configuration for \model-base was RoBERTa lr=1e-5, task lr=1e-4.  All other hyperparameters were the same as in the original implementation.  Training takes about 10 hours on a single GPU.

Our implementation is a superhack that involves PyTorch and Tensorflow sharing a single process and GPU.  To avoid re-implementing the complicated coarse-to-fine logic from Tensorflow in PyTorch (that involves a highly optimized custom GPU kernel originally released by \citet{lee-etal-2018-higher}), we devised a system where the lower transformer portion of the model passes activations and gradients back and forth between PyTorch and Tensorflow.  The input tensors are first run through the transformer in PyTorch, the activations are collected from the top layer, transferred from GPU to CPU then from CPU to Tensorflow and back to GPU to run the coarse-to-fine clustering and compute the loss.  Then gradients are back propogated in Tensorflow to the top of the transformer and the process reversed to transfer them to PyTorch for back propogation through the remainder of the model.  Separate optimizers are maintained with identical LR schedules for parameter updates.  The overhead in this approach is minimal compared to the overall cost of running the model.

\paragraph{Text classification}
For classification, following BERT, we used a simple binary cross entropy loss on top of a first \texttt{[CLS]} token with addition of global attention to \texttt{[CLS]}. We used Adam optimizer with batch sizes of 32 and linear warmup and decay with warmup steps equal to 0.1 of the total training steps. For both IMDB and Hyperpartisan news we did grid search of LRs [3e-5, 5e-5] and epochs [10, 15, 20] and found the model with [3e-5] and epochs 15 to work best. Experiments were done on a single RTX8000 GPU. 



\end{document}